\documentclass[conference]{IEEEtran}
\usepackage{graphicx}
\usepackage{natbib}
\usepackage[nolist]{acronym}
\usepackage[mediumspace,mediumqspace,Grey,squaren]{SIunits}
\usepackage{booktabs}
\advance\topmargin-1in\advance\oddsidemargin-1in
\evensidemargin\oddsidemargin

\makeatletter
\def\@normalsize{\@setsize\normalsize{12pt}\xpt\@xpt
\abovedisplayskip 10pt plus2pt minus5pt\belowdisplayskip \abovedisplayskip
\abovedisplayshortskip \z@ plus3pt\belowdisplayshortskip 6pt plus3pt
minus3pt\let\@listi\@listI}

\def\section{\@startsection {section}{1}{\z@}{20pt plus 2pt minus 2pt}
{8pt plus 2pt minus 2pt}{\centering\normalsize\sc
\edef\@svsec{\thesection.\ }}}
\def\thesection{\Roman{section}}

\def\subsection{\@startsection {subsection}{2}{\z@}{16pt plus 2pt minus 2pt}
{6pt plus 2pt minus 2pt}{\normalsize\sl
\edef\@svsec{\thesubsection.\ }}}
\def\thesubsection{\Alph{subsection}}

\long\def\@makecaption#1#2{
\vskip10pt\begin{center} #1 #2 \end{center}\par\vskip 1pt}
\def\fnum@figure{\raggedright{\footnotesize Fig. \thefigure }.%
\footnotesize}
\def\fnum@table{\footnotesize TABLE \thetable\\\footnotesize\sc}
\def\thetable{\Roman{table}}

\makeatother

\begin{acronym}
  \acro{APE}{absolute percentage error}
  \acro{DE}{differential evolution}
  \acro{EDA}{electronic design automation}
  \acro{IQR}{interquartile range}
  \acro{IC}{integrated circuit}
  \acro{NSGA}{nondominated sorting genetic algorithm}
  \acro{PPA}{power, performance, and area}
  \acro{ROM}{read-only memory}
  \acro{RF}{register file}
  \acro{SoC}{system-on-chip}
  \acro{SRAM}{static random access memory}
  \acro{VT}{voltage threshold}
  \acro{SRAM}{static random access memory}
  \acrodefplural{SRAM}[SRAMs]{static random access memories}
  \acrodefplural{SoC}[SoCs]{systems-on-chip}
\end{acronym}

\begin{document}
\date{}

\title{\Large\textbf{Differentially Evolving Memory Ensembles: \\
Pareto Optimization based on Computational Intelligence for Embedded Memories on a System Level}}

\author{
        Felix Last
        \\ \textit{Technical University of Munich} 
        \\ \textit{and Intel Germany}
         \\ \small{mail@felixlast.de}
    \and 
        Ceren Yeni
        \\ \textit{Intel Germany}
         \\ \small{cerenyeni@gmail.com}
    \and
        Ulf Schlichtmann
        \\ \textit{Technical University of Munich}
        \\ \small{ulf.schlichtmann@tum.de}
}

\maketitle
\thispagestyle{empty}

{\small\textbf{Abstract---
    As the relative power, performance, and area (PPA) impact of embedded memories continues to grow, proper parameterization of each of the thousands of memories on a chip is essential. When the parameters of all memories of a product are optimized together as part of a single system, better trade-offs may be achieved than if the same memories were optimized in isolation. However, challenges such as a sparse solution space, conflicting objectives, and computationally expensive PPA estimation impede the application of common optimization heuristics. We show how the memory system optimization problem can be solved through computational intelligence. We apply a Pareto-based Differential Evolution to ensure unbiased optimization of multiple PPA objectives. To ensure efficient exploration of a sparse solution space, we repair individuals to yield feasible parameterizations. PPA is estimated efficiently in large batches by pre-trained regression neural networks. Our framework enables the system optimization of thousands of memories while keeping a small resource footprint. Evaluating our method on a tractable system, we find that our method finds diverse solutions which exhibit less than 0.5\% distance from known global optima.
}}

\section{Introduction}
        Modern \acp{IC} often contain more than 500 embedded memories. As technology nodes advance to ever smaller transistor lengths, the relative impact of memories on overall circuit \ac{PPA} continues to grow. The multi-objective optimization of memory \ac{PPA} is therefore a crucial component of \ac{EDA} flows. Previous research which focused on optimizing parameters of individual memories for balanced \ac{PPA} fails to regard the \ac{IC}'s memory inventory as a \textit{system}, in which \ac{IC} \ac{PPA} targets may be achieved by trading off \ac{PPA} across different memories. For example, the ability of one high-density memory to achieve exceptionally low area may be exploited by optimizing another memory for low power usage, neglecting area. These trade-offs become possible only when \ac{PPA} of all memories of an \ac{IC} (henceforth: system) are considered as a whole, rather than in isolation.

        The multi-objective \ac{PPA} optimization problem is characterized by the conflicting objectives of power, performance, and area, where the improvement of one often leads to deterioration of the other \cite{Siddavaatam.2019}. Any decision is a trade-off between multiple conflicting objectives as well as any constraints. Without assumptions about the importance of individual objectives, and in the presence of different data scales, multi-objective problems are most commonly solved through Pareto-based methods \cite{Deb.2002}.
        
        When optimizing a single memory, exhaustive search of the design space has been shown to be feasible \cite{Last.2020}. However, the combinatorial explosion of the design space renders such an approach infeasible for the optimization of entire systems. The area of computational intelligence is gaining research attention in many real-world applications of optimization due to these algorithms' high capability to deliver near-optimal solutions for global optimization problems \cite{Goel.2020}. However, their application to the system optimization problem of embedded memories poses some challenges in terms of algorithm selection and refinement as well as problem modeling. In this work:
        \begin{itemize}
            \item we show how the evolutionary algorithm ``differential evolution'' can be used to optimize entire systems of memories for multiple PPA objectives,
            \item we explain the specific challenges that arise in problem modeling, and
            \item we demonstrate that our approach leads to fast optimization which finds near-optimal solutions.
        \end{itemize}

        Our work is structured as follows: In Section \ref{sec:motivation} we discuss and illustrate the motivation for system-level memory \ac{PPA} optimization. Section \ref{sec:problem-definition} defines the problem, for which we outline our solution proposal in Section \ref{sec:proposed-solution}. In Section \ref{sec:results} we present the results of our evaluation of the proposed solution. Finally, we conclude our work in Section \ref{sec:conclusion}, providing an outlook on future research.
        
        \section{Motivation}
        \label{sec:motivation}
\begin{figure}[htb]
        \includegraphics[width=\columnwidth]{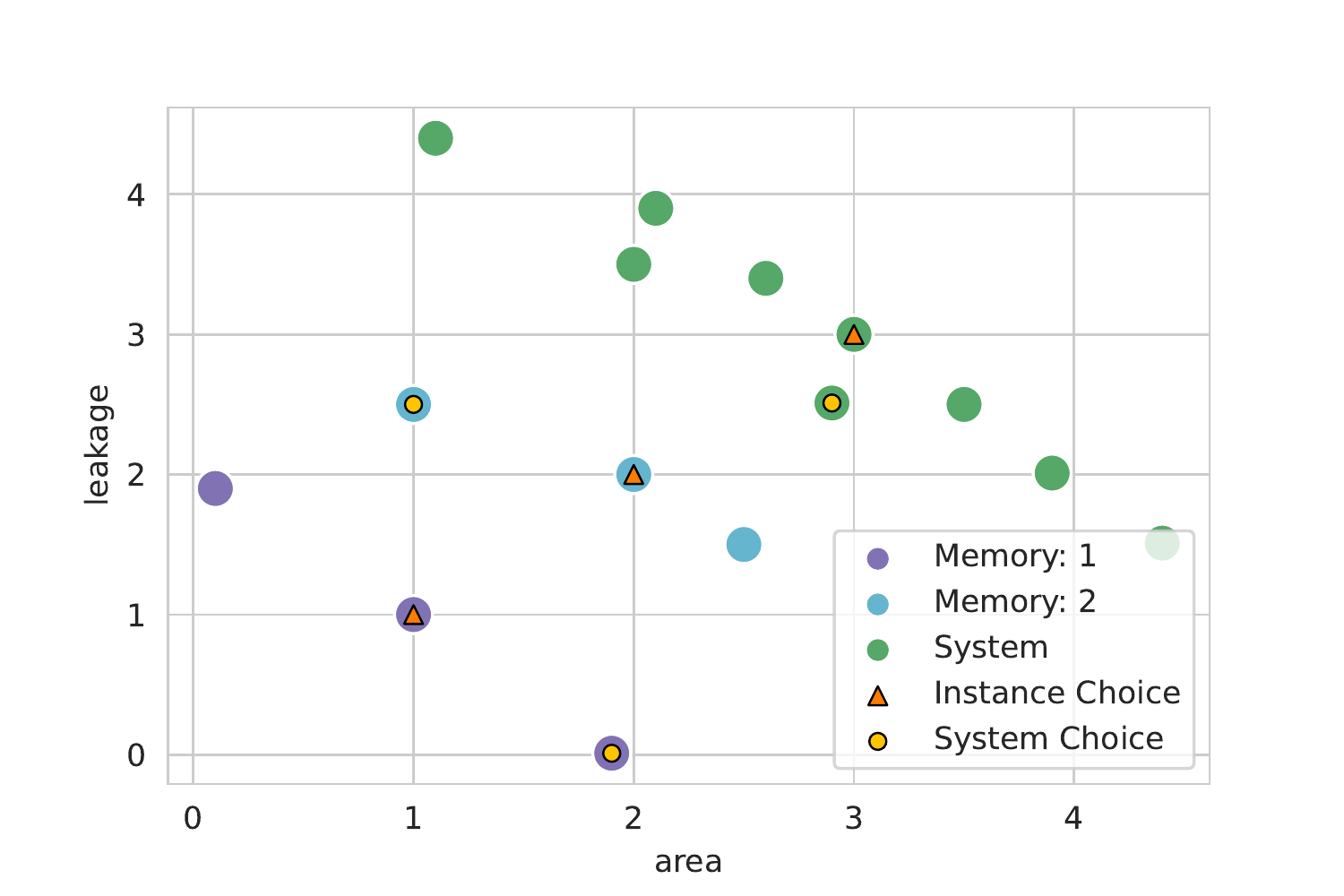}
        \caption{In this toy system, choosing the most balanced trade-off on instance level leads to a sub-optimal choice on system level}
        \label{fig:toy-system}
        \end{figure}

        System optimization may find solutions which objectively outperform those chosen through instance optimization. To understand why better decisions can be made on system level, consider the following toy example: A system to be optimized consists of two memories, where the design space spans three solution candidates per memory. The goal is to achieve a good trade-off between area and leakage, with equal importance given to both objectives, which are assumed to be in the same scale. The three candidates of the first memory have an area and leakage of $(1.0, 1.0), (1.9, 0.01), (0.1, 1.9)$. For the second memory, the candidates are $(2.0, 2.0), (1.0, 2.5), (2.5, 1.5)$. The candidates are shown in Figure \ref{fig:toy-system} as ``Memory 1'' and ``Memory 2''.
        
        For each memory, there is one candidate with equal leakage and area and two candidates which trade off one objective at the cost of the other. In instance-based optimization, the memories are optimized separately. When the goal is to achieve balanced product \ac{PPA}, i.e. equal importance for area and leakage, most designers would choose the center option for memory 1 and 2 which has equal area and leakage (marked as ``Instance Choice'').
        
        When considering the system as a whole, there are 9 solution candidates, which are the combination of the $3 \times 3$ solution candidates. For each combination, area and leakage are summed; the results are shown in Figure \ref{fig:toy-system} as ``System''. The individually balanced choice made above is marked as ``Instance Choice''. Below that choice, there is another solution candidate ''System Choice'', which objectively outperforms the instance-based choice in both objectives. Therefore, when considering the system as a whole, no rational designer would choose the option they would have chosen in instance-based optimization. Also note that the combination of two choices which are Pareto-optimal (nondominated) on an instance level turns out to not be Pareto-optimal on a system level. It is therefore highly desirable to optimize memories not individually, but in combination when their combined \ac{PPA} (i.e. the overall product \ac{PPA}) are the true objectives.
        
        Another benefit of system optimization is that it involves less work for the design team. Choosing a solution from the Pareto set is not trivial. Although techniques for aiding the selection have been proposed (e.g. algorithms for finding the knee point \cite{Satopaa.2011}), this step is rarely fully automated as it involves expert preference. In instance optimization, a choice has to be made for each memory, whereas in system optimization only a single choice is made. Moreover, products' \ac{PPA} targets often change over time as e.g. power becomes more important in later design stages. In that case, system optimization allows designers or even architects to quickly adjust all of the product's memories to meet their new targets.

\section{Problem Definition}
\label{sec:problem-definition}
    \subsection{Memory Compilers}
        The large number of memories on modern \acp{IC} (often between 500 and 10,000) and the regular structure of their bit cell arrays lead \ac{EDA} vendors to create memory compilers which automate the generation of memory circuits. For a single technology node, multiple compilers may exist. They differ in which bit cell type is used, what type of memory they produce (e.g. \acp{SRAM}, \acp{ROM}, \acp{RF}) and the number of I/O ports their memories exhibit. The choice of bit cell also implies a specific \ac{PPA} focus, such as area (high density) or performance (high speed).
        
        When generating memories through memory compilers, system and architectural parameters need to be specified. While system parameters define the memory's interface and result directly from the designer's needs (e.g. number of words, word length, number of ports), architectural parameters affect the \ac{PPA} of the resulting memory, but not it's interface. Examples of architectural parameters include the number of banks, number of column multiplexers. The goal in memory optimization is therefore to select values for architectural parameters to the end of achieving optimal \ac{PPA}, a task frequently referred to as ``knob tuning''. Memory compilers offer an ever-growing number of options, increasing the complexity of the optimization problem. In our work, we use memory compilers with around ten architectural parameters to be tuned.
        
        Another parameter to be optimized is the compiler choice itself. System parameters often allow different compilers to be used. For example, multiple single-port compilers may exist, leaving the choice of memory type and \ac{PPA} focus up to the designer. Because different compilers offer different parameters, having to simultaneously optimize compiler choice and compiler parameterization adds to the complexity of problem modeling.
        
        Our problem is characterized by multiple objectives. The specific objectives differ between products, but may involve area, aspect ratio, access and cycle time, as well as static and dynamic power consumption. Timing and power are evaluated at a specified process, voltage, and temperature. Area, power, timing and aspect ratio are measured in different units and all objectives exhibit a different scale. Like \cite{Last.2020}, we utilize previously trained artificial neural networks to rapidly estimate these objectives for any given compiler parameterization instead of using the compiler directly, which would be very expensive in terms of time and computational resources.

    \subsection{Feasibility Constraints}
        Naturally, the design space posed by a memory compiler's parameters is constrained because not all parameterizations are feasible. We observe two types of feasibility constraints which shape the design space: 1) The set of values which a given parameter can take on is discrete and finite. For example, the threshold voltage of the periphery ($V_{th}$) can take on any of $\{\mathit{low-vt}, \mathit{standard-vt}, \mathit{high-vt}\}$, posing a ``choice constraint''. 2) Based on the number of words and number of bits, some parameter groups are constrained to certain combinations, posing a ``combinatorial constraint''. For example, large memories require large column multiplexing ratios and a higher number of banks (and vice-versa). Consequently, only the following combinations may be allowed for the column multiplexing ratio and number of banks of a certain memory: $\{(4,4), (4,8)\}$. 
        
        As a consequence of feasibility constraints, the design space is sparse. This poses a challenge for many optimization algorithms which commonly rely on random changes to a solution candidate in order to explore the design space. Random changes to compiler parameters are not guaranteed, and in fact unlikely, to produce a feasible solution.
        
        The dimensionality of the design space is equal to the number of parameters of a solution. In instance optimization, this length is equal to the number of architectural parameters plus one parameter for the compiler choice. In system optimization, on the other hand, one solution consists of all memories' architectural parameters, plus the compiler choice for each memory. For example, for a product with 1,000 memories and 9 architectural parameters per memory, the design space has 10,000 dimensions.

\section{Proposed Solution}
\label{sec:proposed-solution}
    \subsection{Algorithm}
        According to the ``no free lunch'' theorems \cite{Wolpert.1997} no single optimization algorithm outperforms all other algorithms on all problems. Instead, some algorithms are better suited to some class of problems. Therefore, the choice of algorithm must be made in accordance with problem characteristics. In this section, we describe the problem characteristics which informed our algorithm choice and explain the chosen algorithm's functioning.
        
        The memory system optimization problem is characterized by having multiple, continuous objectives in different scales. Therefore, multiple optimal solutions exist in the Pareto sense and several should be returned for the designer to choose from. The objective function is non-linear and non-convex. The parameters to be optimized are discrete valued and very numerous due to the combinatorial nature of the problem. Convergence to a global optimum in a short time is desirable. The optimization should be robust, stable, and scalable and provide thorough design space exploration. Parallelizability and low resource usage are desirable, but not essential.
        
        To inform our decision, we compare these requirements to various optimization algorithms' properties based on a thorough literature review. We adapt the ``traceability matrix'' technique from software development to the systematic selection of algorithms. To this end, we create a table which maps requirements (in the table head) to algorithms (in the first column). Each table cell is then the intersection of an algorithm with a requirement. If the algorithm fulfills the requirement according to literature, we set the cell to one, otherwise to zero. Finally, we score each algorithm by computing a weighted sum of each row, using indicators of importance assigned to each requirement as weights. Any algorithms which don't meet key requirements are discarded. We compare 40 computational intelligence algorithms found in \cite{Goel.2020}. The highest-weighted criteria (weighted 3) are multi-objective optimization, handling of large solution vectors, scalability, parellelizability, and fast convergence. Through this method, the following five algorithms are found to be most suitable (score in parentheses): \ac{DE} (25), Artificial bee colony (18), Ant colony optimization (15), Bat algorithm (14), and Artificial fish swarm (13). Hence, we select \ac{DE}, which is described in the following.
        
        Differential evolution \citep{Storn.1997} belongs to the family of evolutionary algorithms, which is a branch of nature-inspired computational intelligence algorithms \citep{Goel.2020}. Like genetic algorithms, it works by evolving a population of individuals over several generations. Each generation consists of two phases: crossover/mutation and selection. For crossover and mutation, a special differential operator is used, which may be parallelized for improving performance. The differential operator combines two random individuals into a mutant vector before stochastically combining it with another parent individual to create an offspring. Previous research \citep{Storn.1997,Das.2011} has found that \ac{DE} converges faster and more reliably than other methods, and its small number of hyperparameters (crossover probability $\mathit{CR}$ and differential weight $F$) make it attractive in practice. Notably, it works on non-differentiable, nonlinear and multimodal objective functions.
        
        In the selection step, the objective function is evaluated for the entire population of parents and offspring. The original algorithm is designed for single-objective optimization and greedily retains the better half of individuals. Because we require Pareto optimization, this step of the algorithm is modified. \cite{Madavan.2002} have proposed to use \ac{DE} in conjunction with the selection procedure of \ac{NSGA}-II. \ac{NSGA}-II was proposed by \cite{Deb.2002} and works by sorting individuals according to their nondomination rank, where the first rank is the Pareto front. In that sense, the method is elitist as it preserves the best individuals. Moreover, a crowding distance is employed to encourage diversity in the last selected nondomination rank \cite{Madavan.2002}.
        
    \subsection{Problem Modeling}
        The successful application of computational intelligence hinges on suitable problem modeling. Modeling the problem of system-level memory optimization poses several unique challenges which we address in this section. Note that although we focus on modeling appropriately for the chosen \ac{DE} algorithm, the same model could be employed in conjunction with many other methods.
        
        In evolutionary algorithms, potential solutions are modeled as vectors which are referred to as individuals of a population. To model solutions to the system optimization problem, we concatenate the solution vectors of all memories. One individual therefore represents the parametrization of all memories of a system. As such, individual vectors are very big, with 5,000 to 10,000 elements. \ac{DE} evolves a whole population of these individuals, where common population sizes range from four to thirty individuals.
        
        Choosing which memory compiler to use to generate a given memory lays the foundation for parameter tuning, as it controls the overall \ac{PPA} focus. The compiler choice is also a core modeling challenge, because it determines which parameters are available and which values are feasible. However, the size of an individual is assumed to be constant in \ac{DE}. To determine the individual vector length, we firstly determine which compilers correspond to the number of I/O ports required by each memory. For each memory, we then allocate as many vector elements as the number of architectural parameters of that compiler with the most parameters.
        
        As discussed in Section \ref{sec:problem-definition}, two types of solution constraints exist: choice constraints, i.e. which values are permitted for individual parameters, and combinatorial constraints, i.e. which combinations of parameters are permitted. Both types of constraints also depend on the compiler choice. The resulting design space is discrete and sparse. This poses a problem for \ac{DE} which, in its original formulation, mutates real-valued individuals semi-randomly. Therefore, \ac{DE} mutation produces almost exclusively infeasible solutions.
        
        The most popular approach for dealing with feasibility constraints in computational intelligence is to include a penalty term for infeasible solutions in the objective function \cite{Orvosh.1994,Deb.2002}. However, evaluating the objective functions for infeasible memories is not possible. Additionally, Pareto-based multi-objective evaluation (as in \ac{NSGA}-II selection) does not exhibit a single objective value to which the penalty term would be added. Differences in scale further exacerbate choosing an adequate penalty parameter. 
        
        Another approach to constraint handling involves discarding infeasible solutions. Due to the sparsity of the design space, too many solutions would be discarded even if the algorithm produced discrete valued individuals. Lastly, the algorithm itself could be modified to produce only feasible solutions, e.g. by introducing specialized mutation and cross-over operators \cite{Orvosh.1994}. However, calculating feasible mutations is computationally demanding due to parameter interactions and would significantly increase run time.
        
\begin{figure}[htb]
        \includegraphics[width=\columnwidth]{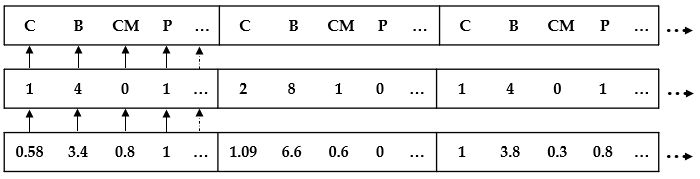}
        \caption{Repairing an unfeasible, real-valued individual vector to make it feasible}
        \label{fig:solution-vector-repair}
        \end{figure}

        To meet feasibility constraints efficiently, we adopt a "repair" strategy \cite{Orvosh.1994}. Before evaluating the objective function, any individual which violates constraints is converted into one which does not. Figure \ref{fig:solution-vector-repair} illustrates how a real-valued solution vector is converted to a discrete-valued, feasible parameterization
        
        The repair is performed in three steps: Firstly, we repair the compiler choice parameter, secondly, we repair all choice-constrained parameters and lastly, we repair combinatorially constrained parameters.
        
        For the first two repair steps, we construct a look-up table for each parameter (see example in Table \ref{table:lookup}) which maps feasible, non-numerical parameter values to discrete integers. For a given real value from the solution vector, we find the closest integer from the look-up table (last column) and select the respective parameter value (second column). This look-up is performed for the compiler choice first, as the compiler is required for the subsequent parameter look-ups.
        
        \begin{table}[!tb]
            \caption{Lookup tables are used to map continuous values of evolved individuals to feasible compiler parameter choices}
            \begin{minipage}{\columnwidth}
            \centering
            \def\arraystretch{1.5}\tabcolsep 6pt
            \def\thefootnote{a}\footnotesize
            \begin{tabular}{l@{~~~~~}l@{~~~}l}
            \hline
            \parbox[c]{7mm} {\textbf{Compiler}} & {\textbf{$V_{th}$}} & \textbf{$\overline{V_{th}}$}\\
            \hline
             \hline
             compiler A &low-vt & 0 \\ 
             \hline
             compiler A &standard-vt & 1  \\
             \hline
             compiler A &high-vt & 2  \\  
             \hline
             compiler B &low-vt & 0 \\
             \hline
             \ldots & \ldots & \ldots \\ 
             \hline
            \end{tabular}
            \end{minipage}
            \label{table:lookup}
        \end{table}

        For combinatorial constraints, we firstly determine all feasible combinations for the given compiler choice. For each feasible combination, we compute the pair-wise Euclidean distance to the corresponding real values from the solution vector. The closest feasible combination is chosen according to the lowest distance.
        
        Once an individual has been ``repaired'' to be feasible, the objective function can be computed. The repair strategy is only used to enable objective function computation. After the selection stage, individuals are not replaced \cite{Orvosh.1994} by their repaired counterparts, retaining population diversity in the otherwise sparse design space.
        
        The objective function is estimated through pre-trained artificial neural networks. Since neural network estimates are matrix computations, optimal performance is achieved when predictions are made in batches. After generating and repairing all offspring, we extract the individual memories' parameterizations and group them across individuals with other memories of the same compiler. This allows us to make a single batch prediction per compiler. Afterwards, \ac{PPA} values need to be regrouped again, so that they can be aggregated by individual to serve as the objective function value.
        
\section{Results}
\label{sec:results}
    In this section we present our evaluation methodology and assess the proposed solution on a small, tractable system, comparing it to an exhaustive search baseline.

    \subsection{Methodology}
        
        We created two sets of sample memory requirements, which we refer to as example systems. Both example systems consist of four memories, for each of which the design space spans between 50 and 900 solution candidates (200 on average). The combined design space (i.e. the system-level design space) consists of approximately 500 million solution candidates (the product of the number of solution candidates on instance level), for which around 1,000 unique \ac{PPA} evaluations are required (the sum of the number of unique solution candidates on instance level).
        The systems are chosen to be small on purpose to allow comparison to known global optima, which can be found through exhaustive search.
        
        Objective evaluation of multi-objective optimization results is not trivial \cite{Li.2020}. This is in part due to the fact that Pareto-optimization doesn't yield single solutions, but sets of equally good solutions. An intuitive approach is to report distribution parameters of the Pareto set, such as the mean and extreme points of each objective. As pointed out by \cite{Li.2020}, distribution parameters (descriptive statistics) alone may be misleading as they don't capture all characteristics decision makers may find important, such as solution diversity. Therefore, we additionally perform visual inspection of results, the most common evaluation technique for Pareto optimization results. We focus on two objectives (area and power) in this work so that visual evaluation is straightforward. Finally, we also measure how expensive the optimization is in terms of time and computational resources.
        
        The proposed algorithm (\ac{DE} with \ac{NSGA}-II selection) is applied to each example systems three times. Every repetition uses a different initial random population. The population consists of 20 individuals which are evolved over 50 generations. The method's hyperparameters are set to $\mathit{CR}=0.9$ and $F=0.8$, respectively. These choices are based on practitioners' rule of thumb, not on fine-tuning.
        
        Our baseline method works as follows: We gather \ac{PPA} estimates exhaustively for all candidates of each memory. We then sum the \ac{PPA} for all possible combinations, i.e. all 500 million solutions of the system-level design space. Finally, the Pareto set is extracted using a divide-and-conquer method \cite{Borzsony.2001}, which is also used to extract the Pareto set after the last generation of \ac{DE}.
        
        \subsection{Method Evaluation}
\begin{figure}[htb]
            \includegraphics[width=\columnwidth]{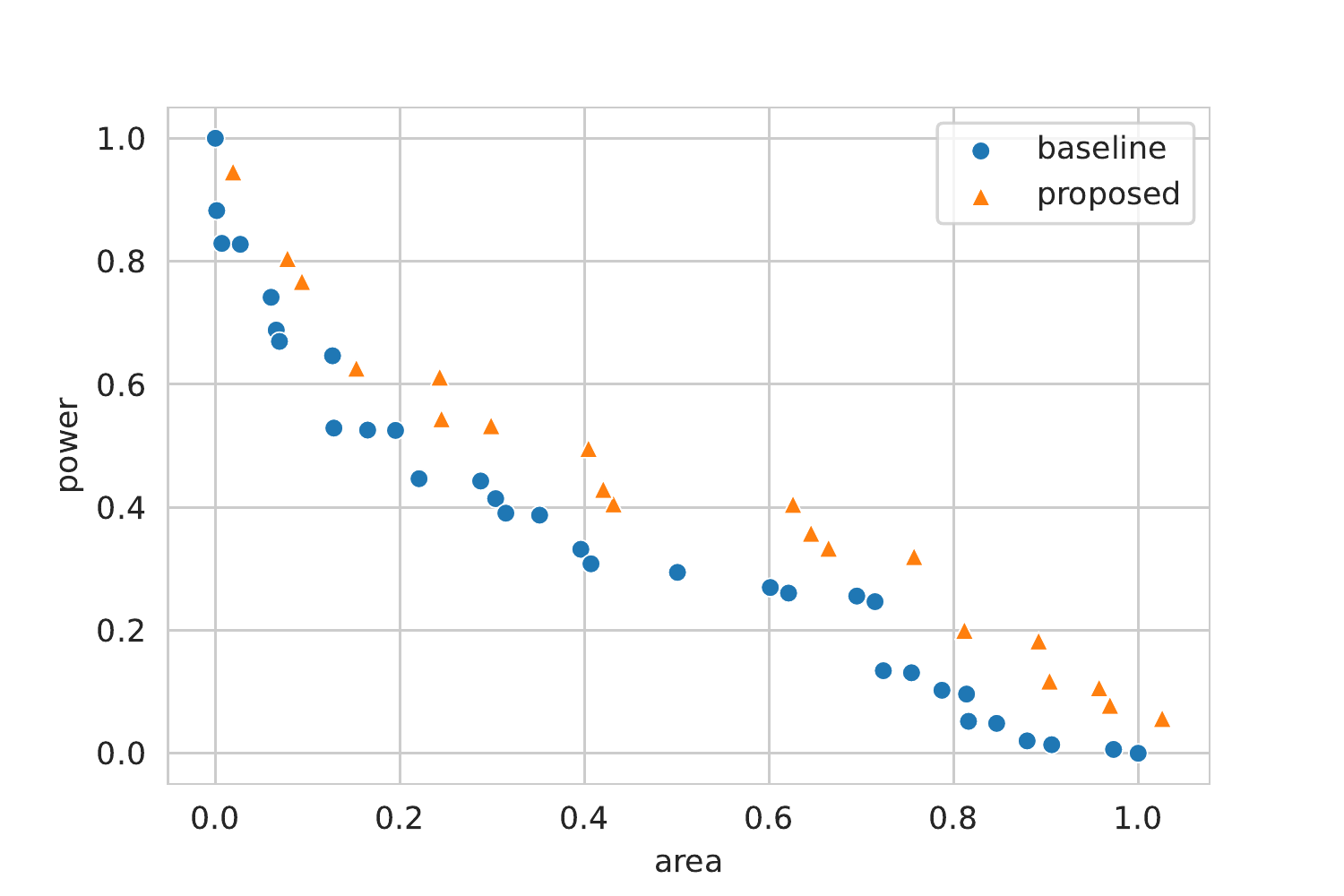}
            \caption{System A: Pareto fronts found by proposed method and global Pareto set}
            \label{fig:proposed-vs-baseline-system-a}
            \end{figure}

            Figure \ref{fig:proposed-vs-baseline-system-a} presents the results of our experiments conducted on example system A. The figure shows the true, global Pareto set found by exhaustive search as ``baseline'', and the front found by one repetition of the proposed method as ``proposed''.  The axes for area (x) and power (y) are normalized by the baseline's minima and maxima. The Pareto set found through exhaustive search shows a dense line of solutions which follow the expected trade-off between the conflicting objectives, area and power. Results obtained by the proposed method closely resemble the shape of the global Pareto set, albeit slightly worse. Although the solution density is lower for the the proposed method, solutions are available throughout each region (low area, low power, balanced). 
            
\begin{figure}[htb]
            \includegraphics[width=\columnwidth]{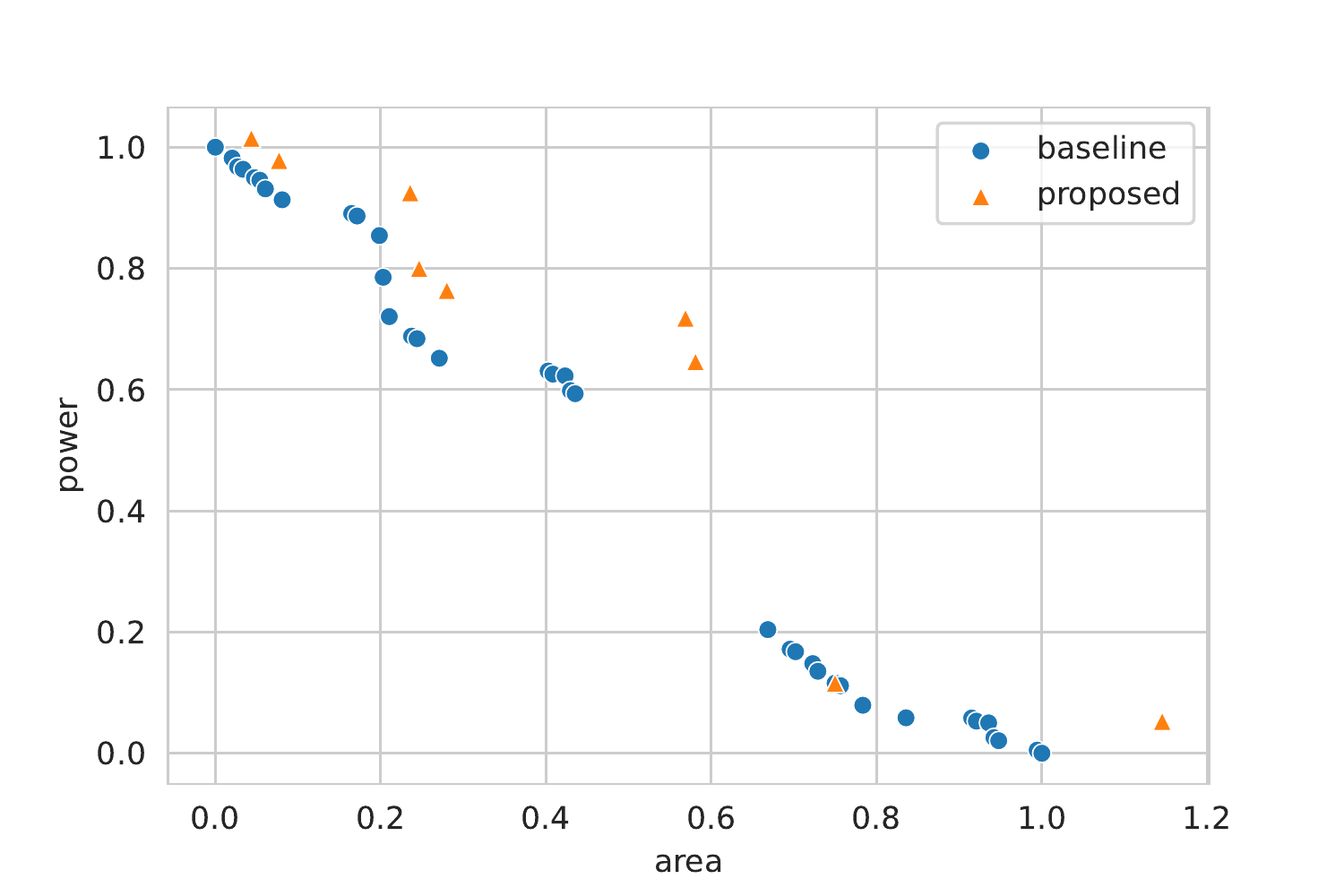}
            \caption{System B: Pareto fronts found by proposed method and global Pareto set}
            \label{fig:proposed-vs-baseline-system-b}
            \end{figure}

            Figure \ref{fig:proposed-vs-baseline-system-b} shows the equivalent results for system B. Here, we observe a less curved Pareto set with a gap in the middle. The proposed method again resembles this shape, with seven points close to the wider, upper segment and two in the lower segment. In the lower segment, we also observe that one point found by the proposed method lies directly on the global Pareto manifold. As in system A, the diversity of the Pareto set is well covered, with extreme points on both ends of the area-power trade-off, and some covering more balanced choices.
            
            To assess how far the solutions found by the proposed method are in relation to their true value, we calculate the deviation of distribution parameters of the Pareto set found by the proposed method from the distributional parameters of the global Pareto set found through exhaustive search. We present these values as a percentage of the latter. For example, if the minimal (best) area attained by our method was 110, while the known best attainable area was 100, then the percentage deviation would be given by $(100-110)/100 = 10\%$. The best area found by our method would therefore be 10\% higher than the globally best area. Table \ref{table:results-system-a} shows these results for example system A. Each row of the table refers to a different distribution parameter, such as the minimum in the previous example. The columns divide these into area and power. Because the proposed method was repeated three times, we report the mean and standard deviation of these values across three repetitions (see second level columns).  For example, the row ``min'' at column ``area deviation mean'' refers to the mean deviation of the minimum area of the solution set found by the proposed method from the minimum area of the global Pareto set. Distributional parameters Q$_1$, Q$_2$, and Q$_3$ refer to the 25$^{th}$ percentile, median, and 75$^{th}$ percentile, respectively.

            \begin{table}[tb]
                \centering
                \caption{System A: Deviation of distribution parameters of Pareto set found by proposed method from global Pareto set (as percentage of the latter), with mean and standard deviation across three repetitions}
                \label{table:results-system-a}
                \begin{tabular}{lllll}
                \toprule
                Pareto set & \multicolumn{2}{l}{area deviation} & \multicolumn{2}{l}{power deviation} \\
                distribution &           mean &      SD &            mean &      SD \\
                \midrule
                count &        -50.51\% &  12.25\% &         -50.51\% &  12.25\% \\
                mean  &          0.69\% &   0.27\% &           0.19\% &   0.31\% \\
                SD    &          -4.5\% &   6.11\% &         -15.36\% &  14.32\% \\
                min   &          0.54\% &   0.37\% &           0.75\% &   0.31\% \\
                Q$_1$ &          0.91\% &   0.44\% &            0.3\% &   0.14\% \\
                Q$_2$ &           1.0\% &    0.2\% &           0.53\% &   0.29\% \\
                Q$_3$ &          0.46\% &   0.24\% &           0.15\% &   0.47\% \\
                max   &          0.03\% &   0.13\% &          -0.97\% &   0.73\% \\
                \bottomrule
                \end{tabular}
            \end{table}
            
            In the first row, ``count'', we find that on average, the proposed solution found 50\% fewer solutions than are known to exist in the Pareto set. On average, the found set's mean area was 0.69\% larger than the global set's mean area, while the set's mean power was 0.19\% higher than the true mean. Comparing the best solutions of each set (row ``min''), the proposed method found memory systems which were, on average, 0.54\% larger and consumed 0.75\% more power than the best attainable. Overall, most distribution statistics shifted by less than 1\% on average with low variance across repetitions (standard deviation less than 0.8\%).
            
            Table \ref{table:results-system-b} shows the same analysis conducted for example system B.

            \begin{table}[tb]
                \centering
                \caption{System B: Deviation of distribution parameters of Pareto set found by proposed method from global Pareto set (as percentage of the latter), with mean and standard deviation across three repetitions}
                \label{table:results-system-b}
                \begin{tabular}{lllll}
                \toprule
                Pareto set & \multicolumn{2}{l}{area deviation} & \multicolumn{2}{l}{power deviation} \\
                distribution &           mean &      SD &            mean &     SD \\
                \midrule
                count &        -81.98\% &   5.63\% &         -81.98\% &  5.63\% \\
                mean  &         -0.07\% &   0.11\% &           0.45\% &  0.43\% \\
                SD    &         -8.59\% &  12.42\% &          -1.23\% &  5.47\% \\
                min   &          0.32\% &   0.06\% &           0.28\% &  0.15\% \\
                Q$_1$ &          0.78\% &   0.73\% &           1.22\% &  0.99\% \\
                Q$_2$ &         -0.01\% &   0.71\% &           0.24\% &  0.63\% \\
                Q$_3$ &         -0.77\% &   0.62\% &           0.06\% &  0.18\% \\
                max   &         -0.47\% &   1.11\% &           0.08\% &   0.3\% \\
                \bottomrule
                \end{tabular}
            \end{table}
            
            As with system A, the best solutions found for system B by the proposed method are very close to the best attainable area and power. The best found solutions are only 0.32\% larger and consumed 0.28\% more power. Solutions close to these extreme points of the Pareto curve are found consistently across repetitions as indicated by the low standard deviation for both measures. The number of found solutions, on the other hand, is much lower (-82\%) as compared to the known number of solutions. However, as the visual analysis in Figure \ref{fig:proposed-vs-baseline-system-b} reveals, these solutions are still well diversified. Even in repetitions which found only four solutions, these were spread across both ends of the objective trade-off.

            The differences between proposed method and exhaustive search in terms of resource requirements are substantial: even for a small system with only four memories, exhaustive search uses approximately 100 GB of working memory to compute the combined system PPAs for all 500 million candidate solutions which takes almost 20 minutes. It is easy to see that exhaustive search is not scalable for larger systems, where the number of solution candidates increases exponentially. The proposed method, on the other hand, computes the system PPA of only 40 candidate solutions per generation, using negligible amounts of working memory. The run time of the proposed solution averaged to less than nine minutes (standard deviation: 77 seconds) across three repetitions on the example system A. The proposed method's implementation has not been optimized for run time; both methods include variable network latency for querying PPA estimates. Note that run times are reported for example system B.

    \subsection{Effect of Hyperparameters}
        To assess the method's sensitivity to the choice of hyperparameters, we conducted 9 different experiments varying $F$ and $\mathit{CR}$, as well as six experiments for the population size and number of generations. The system used for these experiments is system B.
        
        For the first set of experiments, we evaluate a grid of $F={0.4, 0.8, 2.1}$ and $\mathit{CR}={0.5, 0.7, 0.9}$. Across each combination, we hold the population size and number of generations constant at 10 and 20, respectively. Our findings are as follows: the best mean area (167.99) as well as the best mean power (4.29) is attained by $F=1.2$ and $\mathit{CR}=0.5$. This configuration also achieves the best minima and maxima for both area and power. The worst configuration in terms of area mean ($174.11$) is $F=0.4, \mathit{CR}=0.9$, whereas the worst power mean ($4.40$) is attained by $F=1.2, \mathit{CR}=0.9$. Overall, a lower $\mathit{CR}$ is preferred. For the choice of $F$, there is no clear preference from our experiments.
        
        In the second set of experiments, $F$ and $\mathit{CR}$ were fixed to the best configuration found above ($1.2$ and $0.5$). We then conduct optimization runs across a grid of population sizes ${10, 20, 50}$ and number of generations ${20, 50}$. We find that 50 generations consistently outperform 20 generations, which implies that 20 generations are insufficient for convergence. For good mean and worst-case area and power, populations of 10 individuals yield the best results. However, increasing the population size to 50 individuals leads to lower minima for area and power. Generally, a larger number of both generations and individuals is preferred, but even with only ten individuals we attain satisfactory results.
    
\section{Conclusion and Outlook}
\label{sec:conclusion}
       
        In this work, we presented and defined the problem of optimizing \ac{PPA} for all embedded memories of an \ac{IC}. We have shown how the alternative of optimizing each memory in isolation may lead to sub-optimal solutions on system level. The system optimization problem is characterized by a sparse design space shaped by heavily constrained compiler ranges and combinatorial parameter constraints. We proposed that ``repairing'' infeasible solutions by mapping them to the closest feasible solution is the best way of approaching this challenge, and enables solving the problem with metaheuristic algorithms.
        
        Through the application of \acf{DE} paired with \ac{NSGA}-II selection for multi-objective Pareto optimization, we have enabled the previously intractable problem of system-level \ac{PPA} optimization of memoris. The method can be applied to arbitrary systems and does not make assumptions about the scale or importance of the different \ac{PPA} objectives. Testing the method on a small example system, where the global optimum can be assessed, \ac{DE} finds solutions with only 0.32\% larger area, and 0.28\% more power consumption than the global optima (example system B).
        
        Our proposed solution uses less than 1 GB of working memory and less than half the time of exhaustive search. Exhaustive search, on the other hand, is entirely infeasible for larger systems. Even with just four memories, exhaustive search takes up 100 GB of working memory and more than 20 minutes to calculate the \ac{PPA} of the approximately 500 million system states.
        
        While design space constraints have been thoroughly addressed in this work, hard constraints on the objective space (i.e. \ac{PPA}) have not yet been covered. This is important for objectives such as performance, which typically does not need to be maximal, but rather meet the frequency of the circuit. Designers expect similar constraints for the aspect ratio, which is desired to be within certain bounds of squareness. Future work should therefore investigate how hard constraints on objectives can be respected while retaining good design space exploration.
        
        Another task of future research remains the application of different algorithms. While we systematically chose \ac{DE} based on the specific problem's characteristics, the large number of metaheuristic algorithms proposed in the literature leaves room for exploring alternative approaches.

\bibliographystyle{IEEEtranN}
\footnotesize\bibliography{system-optimization-asp-dac}

\end{document}